\documentclass[runningheads]{llncs}
\usepackage{graphicx}%
\usepackage{multirow}%
\usepackage{amsmath,amssymb,amsfonts}%
\usepackage{mathrsfs}%
\usepackage[title]{appendix}%
\usepackage{xcolor}%
\usepackage{hyperref}%
\usepackage{textcomp}%
\usepackage{manyfoot}%
\usepackage{booktabs}%
\usepackage{algorithm}%
\usepackage{algorithmicx}%
\usepackage{algpseudocode}%
\usepackage{listings}%
\usepackage{float}%
\usepackage{subcaption}%
\usepackage{booktabs}%
\usepackage{array}%
\usepackage{tabularx}%

\begin{document}
\title{CAGMamba: Context-Aware Gated Cross-Modal Mamba Network for Multimodal Sentiment Analysis}
\titlerunning{CAGMamba for Multimodal Sentiment Analysis}  

\author{
Minghai Jiao\inst{1}\ \and 
Jing Xiao\inst{1}\thanks{Corresponding authors.} \and
Peng Xiao\inst{2} \and
Ende Zhang\inst{1} \and 
Shuang Kan\inst{1} \and \\
Wenyan Jiang\inst{1} \and 
Jinyao Li\inst{1} \and
Yixian Liu\inst{1} \and
Haidong Xin\inst{1}
}
\authorrunning{M. Jiao et al.}
\institute{
School of Computer Science and Engineering, Northeastern University,\\
Shenyang, China
\and
Xingning Power Supply Bureau, Guangdong Power Grid Co., Ltd.,\\
Meizhou, China
}

\maketitle

\begin{abstract}
Multimodal Sentiment Analysis (MSA) requires effective modeling of cross-modal interactions and contextual dependencies while remaining computationally efficient. 
Existing fusion approaches predominantly rely on Transformer-based cross-modal attention, which incurs quadratic complexity with respect to sequence length and limits scalability. Moreover, contextual information from preceding utterances is often incorporated through concatenation or independent fusion, without explicit temporal modeling that captures sentiment evolution across dialogue turns. 
To address these limitations, we propose CAGMamba, a context-aware gated cross-modal Mamba framework for dialogue-based sentiment analysis.
Specifically, we organize the contextual and the current-utterance features into a temporally ordered binary sequence, which provides Mamba with explicit temporal structure for modeling sentiment evolution. 
To further enable controllable cross-modal integration, we propose a Gated Cross-Modal Mamba Network (GCMN) that integrates cross-modal and unimodal paths via learnable gating to balance information fusion and modality preservation, and is trained with a three-branch multi-task objective over text, audio, and fused predictions.
Experiments on three benchmark datasets demonstrate that CAGMamba achieves state-of-the-art or competitive results across multiple evaluation metrics.
All codes are available at \url{https://github.com/User2024-xj/CAGMamba}.
\end{abstract}

\section{Introduction}\label{sec1}


Multimodal Sentiment Analysis (MSA) aims to infer sentiment from complementary signals such as text, audio, and visual cues. 
Since affective expression is often ambiguous within any single modality, modeling cross-modal interactions is essential for integrating complementary information across modalities~\cite{hazarika2020misa}.
Despite these advances, two key challenges remain.
First, many fusion models rely on Transformer-style cross-modal attention, which incurs quadratic complexity and becomes increasingly expensive as sequence length grows. 
Second, contextual information is still underutilized in many settings. 
In dialogue-based sentiment analysis, the sentiment of the current utterance often depends on preceding utterances, yet contextual cues are frequently treated as auxiliary features rather than explicitly modeled as ordered inputs for sequence modeling.

Recent work partially addresses these issues. 
MMML~\cite{wu2023mmml} demonstrates the usefulness of contextual information for sentiment prediction, but mainly incorporates context through concatenation or independent fusion, without providing a clear temporal structure for sequence modeling. 
MSAmba~\cite{he2025msamba} introduces the Mamba architecture to MSA and improves computational efficiency. However, its cross-modal fusion mechanism lacks explicit gating control, limiting fine-grained modality interaction. Moreover, contextual features are not organized as temporally structured inputs, leaving the influence of preceding utterances on current sentiment insufficiently modeled. 
As a result, the influence of prior utterances on the current sentiment is not modeled as explicitly as it could be.

To address the above limitations, we propose CAGMamba, a context-aware gated cross-modal Mamba framework for dialogue-based multimodal sentiment analysis. The main contributions of this work are summarized as follows:

\begin{enumerate}
    \item We propose a context-aware sequence construction strategy that arranges contextual and current-utterance features in a temporally ordered manner, enabling Mamba's selective state space mechanism to explicitly model directional sentiment transitions across dialogue turns.

    \item We design a Gated Cross-Modal Mamba Network (GCMN) that maintains parallel cross-modal and unimodal processing paths with learnable gating, allowing instance-level adaptive control over the balance between cross-modal fusion and modality-specific preservation. A three-branch multi-task objective is further incorporated to jointly supervise text, audio, and fused predictions.

    \item Extensive experiments on CMU-MOSI, CMU-MOSEI, and CH-SIMS demonstrate that CAGMamba achieves state-of-the-art or competitive performance across multiple metrics, with an Acc-2 of 88.19 and F1 of 90.09 on CMU-MOSI.

\end{enumerate}

\section{Related Work}


Existing MSA methods can be broadly divided into representation-learning and cross-modal interaction approaches. 
Methods such as MISA~\cite{hazarika2020misa}, ConFEDE~\cite{yang2023confede}, and CLGSI~\cite{yang2024clgsi} focus on disentangling, aligning, or refining modality-specific and shared representations. 
Cross-modal interaction models such as MulT~\cite{hu2025multilevel}, MAG-BERT~\cite{rahman2020integrating}, SmartRAN~\cite{guo2024smartran}, and TsAFN~\cite{liu2025tsafn} further model inter-modal dependencies more explicitly. 
These methods achieve strong performance, but many of them rely on Transformer-based fusion and therefore inherit the computational cost of dense attention.
Beyond sentiment analysis, multimodal fusion has also proven effective in tasks such as movie audio description~\cite{ye2024mmad}, further underscoring the importance of cross-modal alignment.

State space models provide an efficient alternative for sequence modeling. 
S4~\cite{gu2022efficiently} demonstrates the value of structured state space representations for long-range dependencies, and Mamba~\cite{gu2024mamba} further introduces data-dependent selective state spaces with linear-time complexity. 
Mamba has been successfully applied to both vision~\cite{zhu2024vision} and time-series modeling~\cite{ahamed2024timemachine}.
MSAmba~\cite{he2025msamba} makes an early attempt to adapt Mamba for multimodal sentiment analysis, demonstrating its potential in this task. However, existing Mamba-based approaches do not incorporate dialogue-level context into sequence construction, nor provide explicit gating for balancing cross-modal fusion and modality preservation.

Contextual information has also been widely studied in affective computing. 
DialogueGCN~\cite{ghosal2019dialoguegcn}, MM-DFN~\cite{hu2022mmdfn}, DAG-based approaches~\cite{shen2021directed}, and CSMF-SPC~\cite{li2024csmfspc} show that historical utterances help disambiguate current sentiment. 
More recent work, such as MMML~\cite{wu2023mmml}, explicitly validates the contribution of context in MSA. 
However, contextual features are often incorporated through concatenation or loosely coupled fusion, rather than being organized as temporally structured inputs for sequence models. 
This leaves room for a more efficient framework that combines context modeling and cross-modal fusion more tightly.

\section{Methodology}\label{sec11}
\begin{figure}[t]
\centering
\includegraphics[width=\linewidth]{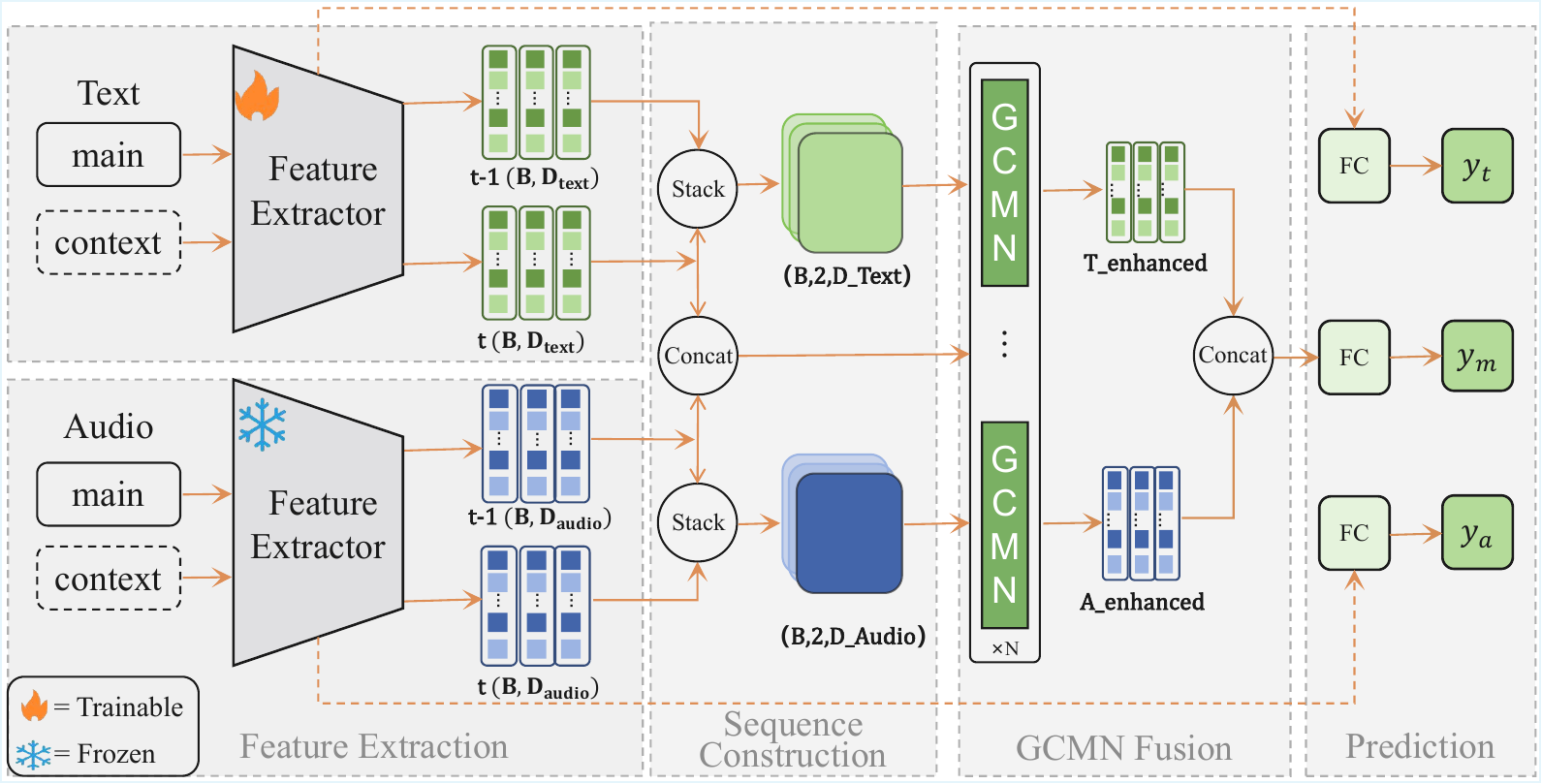}
\caption{Overall Framework of the Proposed CAGMamba Model.}
\label{fig:1}
\end{figure}

In this section, we introduce the proposed \textbf{CAGMamba}
(\textbf{C}ontext-\textbf{A}ware \textbf{G}ated Cross-Modal
\textbf{Mamba} Network), a bimodal sentiment analysis framework
for dialogue-based multimodal sentiment analysis, with its overall
architecture illustrated in Figure~\ref{fig:1}. CAGMamba adopts a
text-audio dual-stream architecture with four stages: feature
extraction, sequence construction, cross-modal fusion, and prediction.

\subsection{Preliminaries}\label{subsec_prelim}

\noindent\textbf{Task Definition.}
Inspired by MMML~\cite{wu2023mmml}, we adopt a text-audio bimodal architecture.
Given a target utterance $u_t$ and its contextual utterance $u_{t-1}$
in a dialogue, each associated with text and audio modalities, the goal
is to predict a sentiment score $y \in \mathbb{R}$ for $u_t$.

\noindent\textbf{State Space Models.}
Mamba maps an input $x(t) \in \mathbb{R}$ to output $y(t) \in \mathbb{R}$
through a latent state $\mathbf{h}(t) \in \mathbb{R}^N$:
\begin{equation}
    \mathbf{h}'(t) = \mathbf{A}\mathbf{h}(t) + \mathbf{B}x(t),
    \quad y(t) = \mathbf{C}^\top \mathbf{h}(t),
\end{equation}
where $\mathbf{A} \in \mathbb{R}^{N \times N}$,
$\mathbf{B} \in \mathbb{R}^{N}$, and $\mathbf{C} \in \mathbb{R}^{N}$
are the state, input, and output matrices. For discrete computation,
the continuous parameters are discretized via zero-order hold with
timescale $\Delta$:
\begin{equation}
    \overline{\mathbf{A}} = \exp(\Delta \mathbf{A}),
    \quad
    \overline{\mathbf{B}} = (\Delta \mathbf{A})^{-1}
    (\exp(\Delta \mathbf{A}) - \mathbf{I}) \cdot \Delta \mathbf{B}.
\end{equation}
Mamba makes $\mathbf{B}$, $\mathbf{C}$, $\Delta$ input-dependent
through learnable projections, enabling selective information filtering
with $O(L)$ complexity.

\subsection{Context-Aware Feature Extraction and Sequence
Construction}\label{subsec_feat_seq}

\noindent\textbf{Feature Extraction.}
A trainable RoBERTa-Large encoder $\mathcal{E}^\text{t}$ and a frozen
Data2Vec-Audio-Large encoder $\mathcal{E}^\text{a}$ extract features
for both the target utterance and its contextual utterance:
\begin{equation}
    \mathbf{F}^\text{t},\; \mathbf{F}^\text{t}_c
    \in \mathbb{R}^{B \times d_\text{t}},
\end{equation}
\begin{equation}
    \mathbf{F}^\text{a},\; \mathbf{F}^\text{a}_c
    \in \mathbb{R}^{B \times d_\text{a}},
\end{equation}
where $d_\text{t}$ and $d_\text{a}$ are the encoder hidden dimensions,
and the $c$ denotes contextual features.

\noindent\textbf{Sequence Construction.}
We project features into a shared fusion space of dimension $f$ via
modality-specific projection matrices $\mathbf{W}^\text{t} \in \mathbb{R}^{d_\text{t} \times f}$
and $\mathbf{W}^\text{a} \in \mathbb{R}^{d_\text{a} \times f}$:
\begin{equation}
    \tilde{\mathbf{F}}^\text{t} = \mathbf{W}^\text{t} \mathbf{F}^\text{t}, 
    \tilde{\mathbf{F}}^\text{a} = \mathbf{W}^\text{a} \mathbf{F}^\text{a},
\end{equation}
\begin{equation}
    \tilde{\mathbf{F}}^\text{t}_c = \mathbf{W}^\text{t} \mathbf{F}^\text{t}_c, 
    \tilde{\mathbf{F}}^\text{a}_c = \mathbf{W}^\text{a} \mathbf{F}^\text{a}_c,
\end{equation}
The projected features are then stacked in
context-to-main temporal order:
\begin{equation}
    \mathbf{S}^\text{t} = [\tilde{\mathbf{F}}^\text{t}_c,\;
    \tilde{\mathbf{F}}^\text{t}] \in \mathbb{R}^{B \times 2 \times f},
\end{equation}
\begin{equation}
    \mathbf{S}^\text{a} = [\tilde{\mathbf{F}}^\text{a}_c,\;
    \tilde{\mathbf{F}}^\text{a}] \in \mathbb{R}^{B \times 2 \times f}.
\end{equation}
This context-to-main ordering allows Mamba's selective scanning to model sentiment evolution directionally: at the
context step, the input-dependent $\overline{\mathbf{B}}_1$ controls
contextual absorption; at the main step,
$\mathbf{h}_2 = \overline{\mathbf{A}}\mathbf{h}_1 +
\overline{\mathbf{B}}_2 \tilde{\mathbf{F}}^\text{t}$
adaptively integrates context with the current input.

\subsection{Gated Cross-Modal Mamba Network (GCMN) Fusion
Module}\label{subsec_gcmn}

\begin{figure}[t]
    \centering
    \includegraphics[width=0.72\textwidth]{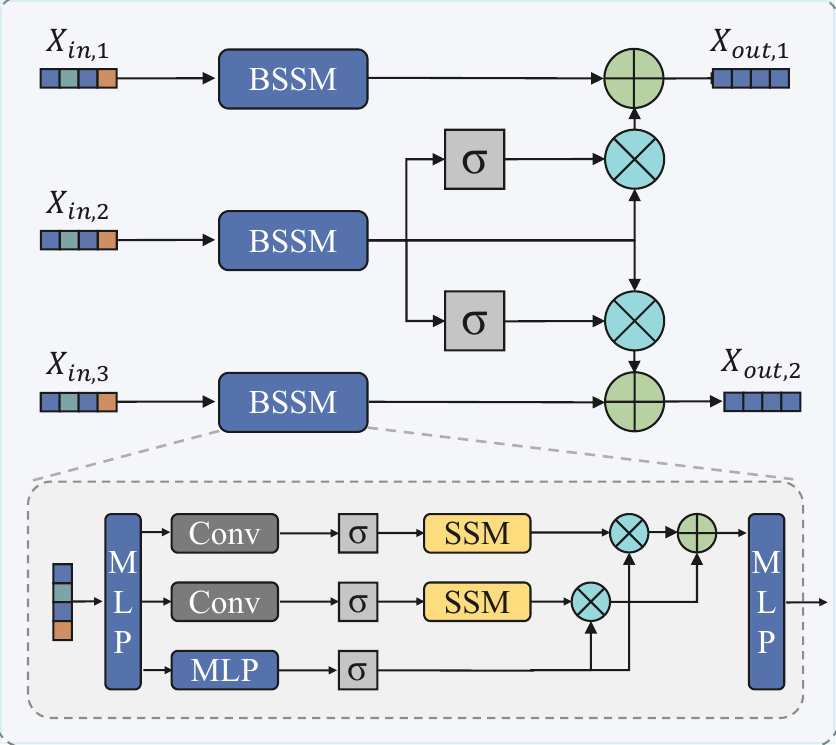}
    \caption{Architecture of the GCMN module. The upper part shows the three-stream gated fusion structure. The lower part illustrates the internal architecture of the BSSM.}
    \label{fig:2}
\end{figure}

GCMN is the core fusion module of CAGMamba, consisting of two components: a Bi-directional Selective Scanning Module (BSSM) for context-sensitive sequence modeling, and a three-stream gated fusion architecture for adaptive cross-modal integration. The overall structure and internal design of BSSM are illustrated in Figure~\ref{fig:2}

\noindent\textbf{Bi-directional Selective Scanning Module.} BSSM combines forward and backward selective scanning with learnable gating to capture bidirectional sequential dependencies, and serves as the fundamental building block for both unimodal and cross-modal processing within GCMN.

Given an input sequence $\mathbf{X} \in \mathbb{R}^{B \times L \times d}$, an MLP first projects it into an expanded hidden space:
\begin{equation}
    \mathbf{Z} = \mathrm{MLP}_\text{in}(\mathbf{X}) \in \mathbb{R}^{B \times L \times d'},
\end{equation}
where $d'$ is the expanded dimension. The projected representation is then split into three parallel branches. Two scanning branches each apply a one-dimensional convolution followed by a SiLU activation:
\begin{equation}
    \mathbf{Z}^\text{fwd} = \sigma_\text{SiLU}(\mathrm{Conv1d}^\text{fwd}(\mathbf{Z})),
\end{equation}
\begin{equation}
    \mathbf{Z}^\text{bwd} = \sigma_\text{SiLU}(\mathrm{Conv1d}^\text{bwd}(\mathrm{flip}(\mathbf{Z}))),
\end{equation}
where $\mathrm{flip}(\cdot)$ reverses the input along the temporal dimension. Each branch is then processed by an independent SSM to model long-range dependencies:
\begin{equation}
    \mathbf{O}^\text{fwd} = \mathrm{SSM}^\text{fwd}(\mathbf{Z}^\text{fwd}),
\end{equation}
\begin{equation}
    \mathbf{O}^\text{bwd} = \mathrm{SSM}^\text{bwd}(\mathbf{Z}^\text{bwd}).
\end{equation}
The outputs of the forward and backward branches are combined via element-wise multiplication to capture bidirectional contextual interactions:
\begin{equation}
    \mathbf{O}^\text{bi} = \mathbf{O}^\text{fwd} \odot \mathrm{flip}(\mathbf{O}^\text{bwd}).
\end{equation}
Meanwhile, a separate gating branch applies an MLP followed by a SiLU activation to produce a gating signal:
\begin{equation}
    \mathbf{G}^\text{ssm} = \sigma_\text{SiLU}(\mathrm{MLP}_\text{gate}(\mathbf{Z})).
\end{equation}
The gating signal modulates the bidirectional output to control the information flow, and the result is projected back to the original dimension through a final MLP:
\begin{equation}
    \mathrm{BSSM}(\mathbf{X}) = \mathrm{MLP}_\text{out}(\mathbf{G}^\text{ssm} \odot \mathbf{O}^\text{bi}).
\end{equation}

\noindent\textbf{Three-Stream Fusion Architecture.}
Each GCMN layer consists of three parallel streams--one cross-modal and two unimodal--with a gating mechanism to balance their contributions.

\noindent\textbf{Cross-Modal Path.}
The text and audio sequences are concatenated along the feature
dimension and processed by a BSSM:
\begin{equation}
    \mathbf{S}^\text{cross} = [\mathbf{S}^\text{a} \| \mathbf{S}^\text{t}]
    \in \mathbb{R}^{B \times 2 \times 2f},
\end{equation}
\begin{equation}
    \mathbf{H}^\text{cross} = \text{BSSM}^\text{cross}(\mathbf{S}^\text{cross}).
\end{equation}
The last time-step output is decoupled into modality-specific
cross-modal features via learned projections
$\mathbf{W}_{\text{c} \to \text{t}},\,
\mathbf{W}_{\text{c} \to \text{a}} \in \mathbb{R}^{2f \times f}$:
\begin{equation}
    \mathbf{F}^\text{t}_\text{cross} =
    \mathbf{W}_{\text{c} \to \text{t}} \cdot \mathbf{H}^\text{cross}[:, -1, :],
\end{equation}
\begin{equation}
    \mathbf{F}^\text{a}_\text{cross} =
    \mathbf{W}_{\text{c} \to \text{a}} \cdot \mathbf{H}^\text{cross}[:, -1, :].
\end{equation}

\noindent\textbf{Unimodal Paths.}
$\mathbf{S}^\text{t}$ and $\mathbf{S}^\text{a}$ are processed
independently by separate BSSMs, yielding unimodal features
$\mathbf{F}^\text{t}_\text{uni}$ and $\mathbf{F}^\text{a}_\text{uni}$
at the last time step.

\noindent\textbf{Gated Fusion.}
The gating weight for each modality is computed from both cross-modal
features jointly:
\begin{equation}
    \mathbf{G}^\text{t} = \sigma\!\big(
    \mathbf{W}^\text{t}_g
    [\mathbf{F}^\text{t}_\text{cross} \| \mathbf{F}^\text{a}_\text{cross}]
    + \mathbf{b}^\text{t}_g\big) \in [0,1]^{f},
\end{equation}
where $\mathbf{W}^\text{t}_g \in \mathbb{R}^{2f \times f}$ and
$\mathbf{b}^\text{t}_g \in \mathbb{R}^{f}$ are learnable parameters;
$\mathbf{G}^\text{a}$ is computed symmetrically. The final
representations are obtained via residual gated fusion:
\begin{equation}
    \mathbf{F}^\text{t}_\text{final} =
    \mathbf{F}^\text{t}_\text{uni} +
    \mathbf{G}^\text{t} \odot \mathbf{F}^\text{t}_\text{cross},
\end{equation}
\begin{equation}
    \mathbf{F}^\text{a}_\text{final} =
    \mathbf{F}^\text{a}_\text{uni} +
    \mathbf{G}^\text{a} \odot \mathbf{F}^\text{a}_\text{cross}.
\end{equation}
GCMN is applied bidirectionally (context$\rightarrow$main and
main$\rightarrow$context) and supports $N$-layer stacking.

\noindent\textbf{Multi-Task Prediction.}
Three prediction heads are employed: two unimodal heads for text and audio, and one multimodal head for the fused representation. The unimodal branches predict from
raw features:
\begin{equation}
    \hat{y}_\text{t} = \mathrm{MLP}_\text{t}([\mathbf{F}^\text{t} \|
    \mathbf{F}^\text{t}_c]),
\end{equation}
\begin{equation}
    \hat{y}_\text{a} = \mathrm{MLP}_\text{a}([\mathbf{F}^\text{a} \|
    \mathbf{F}^\text{a}_c]).
\end{equation}
The fusion branch predicts from all enhanced features:
\begin{equation}
    \hat{y}_\text{m} = \mathrm{MLP}_\text{m}\!\left(
    [\mathbf{F}^\text{t}_\text{final} \| \mathbf{F}^\text{a}_\text{final}
    \| \mathbf{F}^{\text{t},c}_\text{final} \|
    \mathbf{F}^{\text{a},c}_\text{final}]\right).
\end{equation}
The total training loss is:
\begin{equation}
    \mathcal{L} = \alpha \cdot \mathcal{L}_\text{MSE}(\hat{y}_\text{t}, y)
    + \beta \cdot \mathcal{L}_\text{MSE}(\hat{y}_a, y)
    + \gamma \cdot \mathcal{L}_\text{MSE}(\hat{y}_\text{m}, y),
\end{equation}
where $\alpha$, $\beta$, $\gamma$ are branch weights. During inference,
$\hat{y}_\text{m}$ is used as the final prediction.
\section{Experimental Methodology}\label{sec12}
This section describes the datasets, evaluation metrics, baselines, and implementation details used in the experiments.

\textbf{Datasets.}
We evaluate CAGMamba on three benchmarks: CMU-MOSI~\cite{zadeh2016cmumosi}, which contains 2,199 English video segments with continuous sentiment labels; CMU-MOSEI~\cite{zadeh2018cmumosei}, which contains 23,453 samples with a more balanced sentiment distribution; and CH-SIMS~\cite{yu2020chsims}, which provides 2,281 Chinese samples for cross-lingual evaluation.


\textbf{Evaluation Metrics.}
Following prior work~\cite{yang2023confede,yang2024clgsi}, we report binary classification accuracy (Acc-2), F1 score (F1), seven-class classification accuracy (Acc-7), mean absolute error (MAE), and Pearson correlation coefficient (Corr). 
Lower MAE indicates better performance, while higher values are preferred for the remaining metrics. On CH-SIMS, we additionally report Acc-5 and Acc-3.


\textbf{Baselines.}
We compare CAGMamba with representative MSA baselines from three lines of work. 
The first line focuses on representation decomposition and contrastive learning, including MISA~\cite{hazarika2020misa}, Self-MM~\cite{yu2021learning}, MMIM~\cite{han2021improving}, ConFEDE~\cite{yang2023confede}, and CLGSI~\cite{yang2024clgsi}. 
The second line includes recent strong interaction and fusion models, such as Semi-IIN~\cite{lin2024semiiin}, MFMB-Net~\cite{tao2025mfmbnet}, and MISR~\cite{zhu2025misr}. 
The third line contains the methods most related to ours: MMML~\cite{wu2023mmml}, which incorporates contextual information through multi-loss fusion, and MSAmba~\cite{he2025msamba}, which introduces Mamba into MSA for efficient sequence modeling. 
For CH-SIMS, we additionally include LMF~\cite{liu2018efficient}, TFN~\cite{zadeh2017tensor}, MulT~\cite{tsai2019multimodal}, and MAG-BERT~\cite{rahman2020integrating} for broader comparison.


\textbf{Implementation Details.}
For feature extraction, RoBERTa-Large and \\ Data2Vec-Audio-Large are used as the text and audio encoders, both with hidden dimension 1024. 
For the Mamba blocks, the state dimension, convolution kernel size, and expansion factor are set to 16, 4, and 2, respectively. 
We searched the number of GCMN layers over $N \in \{1, 2, 3\}$ and found that $N = 2$ yields the best validation performance; further stacking to three layers leads to slight degradation, likely due to over-smoothing of modality-specific features. The fusion dimension is set to 1024.

The model is trained with AdamW for 200 epochs using batch size 16 and early stopping with patience 8 based on validation loss. 
The multi-task loss weighting coefficient is set to 0.5 across datasets, and a dropout rate of 0.3 is applied.

\section{Experimental Results}\label{sec_results}
We evaluate CAGMamba from four complementary perspectives: comparison with baselines, ablation studies, efficiency analysis, and representation visualization.

\begin{table}[t]
\centering
\caption{Comparison with representative baselines. Acc-2 reports two values: negative/non-negative and negative/positive classification accuracy, respectively. The best results are marked in \textbf{bold}, while the \underline{second-best} results are underlined.}
\label{tab:mosi_mosei}

\resizebox{\linewidth}{!}{
\begin{tabular}{lccccc ccccc}
\toprule
\multirow{2}{*}{\textbf{Method}}
& \multicolumn{5}{c}{\textbf{CMU-MOSI}}
& \multicolumn{5}{c}{\textbf{CMU-MOSEI}} \\
\cmidrule(lr){2-6} \cmidrule(lr){7-11}
& Acc-2 & F1 & Acc-7 & MAE & Corr
& Acc-2 & F1 & Acc-7 & MAE & Corr \\
\midrule

MISA
& 81.80/83.40 & 81.70/83.60 & 42.30 & 0.78 & 0.78
& 83.60/85.50 & 83.80/85.30 & 52.20 & 0.56 & 0.76 \\

Self-MM
& 83.44/85.46 & 83.36/85.43 & 46.67 & 0.71 & 0.80
& 83.76/85.15 & 83.82/84.90 & 53.87 & 0.53 & 0.77 \\

MMIM
& 84.14/86.06 & 84.00/85.98 & 46.65 & 0.70 & 0.80
& 82.24/85.97 & 82.66/85.94 & 54.24 & 0.53 & 0.77 \\

ConFEDE
& 84.17/85.52 & 84.13/85.52 & 42.27 & 0.74 & 0.78
& 81.65/85.82 & 82.17/85.83 & 54.86 & 0.52 & 0.78 \\

CLGSI
& 83.97/86.43 & 83.63/86.25 & 49.67 & 0.70 & 0.79
& 84.01/86.32 & 84.21/86.18 & 54.56 & 0.53 & 0.76 \\

Semi-IIN
& 85.28/87.04 & 85.19/87.00 & 46.50 & 0.68 & 0.82
& 84.98/\textbf{87.70} & 85.27/\underline{87.65} & \textbf{55.89} & \textbf{0.50} & 0.80 \\

MFMB-Net
& 82.70/85.70 & 83.20/86.00 & 45.80 & 0.71 & 0.80
& 84.70/85.10 & 85.00/85.10 & 54.20 & 0.53 & 0.76 \\

MISR
& 86.01/\underline{88.11} & 86.10/\underline{88.15} & 49.85 & \underline{0.67} & 0.82
& 85.28/\underline{87.51} & 85.57/87.55 & \underline{55.05} & \underline{0.51} & 0.79 \\

MMML
& \underline{87.17}/87.15 & \underline{88.01}/88.13 & \underline{51.38} & \textbf{0.56} & \underline{0.86}
& \underline{86.03}/86.12 & \underline{87.81}/\underline{87.65} & 54.37 & 0.53 & \underline{0.81} \\

MSAmba
& 85.99/87.43 & 85.99/87.40 & 49.67 & 0.71 & 0.81
& 85.78/86.86 & 85.99/86.93 & 54.21 & \underline{0.51} & 0.80 \\

\midrule
Ours
& \textbf{88.19}/\textbf{88.15} & \textbf{90.09}/\textbf{90.09} & \textbf{52.48} & \textbf{0.56} & \textbf{0.88}
& \textbf{87.08}/87.14 & \textbf{88.72}/\textbf{88.57} & 54.56 & \underline{0.51} & \textbf{0.82} \\

\bottomrule
\end{tabular}
}
\end{table}

\begin{table}[t]
\centering
\caption{Comparison with representative baselines. The best results are marked in \textbf{bold}, while the \underline{second-best} results are underlined.}
\label{tab:ch_sims_results}

\begin{tabular}{lcccccc}
\toprule
\textbf{Method} & Acc-5$\uparrow$ & Acc-3$\uparrow$ & Acc-2$\uparrow$ & F1$\uparrow$ & MAE$\downarrow$ & Corr$\uparrow$ \\
\midrule

LMF      
& 40.53 & 64.68 & 77.77 & 77.88 & 0.441 & 0.576 \\

TFN      
& 39.30 & 65.12 & 78.38 & 78.62 & 0.432 & 0.591 \\

MulT     
& 37.94 & 64.77 & 78.56 & 79.66 & 0.453 & 0.564 \\

MISA     
& -     & -     & 76.54 & 76.59 & 0.447 & 0.563 \\

MAG-BERT 
& -     & -     & 74.44 & 71.75 & 0.492 & 0.399 \\

Self-MM  
& 41.53 & 65.47 & 80.04 & 80.44 & 0.425 & 0.595 \\

ConFEDE  
& 46.34 & -     & 81.05 & 81.13 & 0.377 & 0.655 \\

MMML     
& \textbf{49.38} & 68.29 & 81.18 & 80.98 & \textbf{0.349} & \underline{0.701} \\

CLGSI    
& 45.95 & -     & 81.18 & 80.59 & 0.408 & 0.634 \\

MSAmba   
& \underline{47.17} & \textbf{68.83} & \underline{82.30} & \underline{81.75} & 0.403 & 0.646 \\

MISR     
& -     & -     & \underline{81.53} & \underline{81.91} & 0.425 & 0.597 \\

\midrule
Ours     
& 47.03 & \underline{68.35} & \textbf{83.13} & \textbf{83.15} & \underline{0.351} & \textbf{0.703} \\

\bottomrule
\end{tabular}
\end{table}

\subsection{Overall Performance}\label{subsec_sota}

Tables~\ref{tab:mosi_mosei} and~\ref{tab:ch_sims_results} summarize the main results on CMU-MOSI, CMU-MOSEI, and CH-SIMS. 
Overall, CAGMamba achieves strong and consistent performance across datasets, suggesting that the proposed architecture generalizes well across languages, data scales, and evaluation protocols.

On CMU-MOSI, CAGMamba achieves the best results on the major metrics, including Acc-2 (88.19/88.15), F1 (90.09/90.09), Acc-7 (52.48), and Corr (0.88), while matching the best MAE (0.56). 
Compared with MMML, which also leverages contextual information, our model improves Acc-2 by 1.02/1.00 points and F1 by 2.08/1.96 points. 
This gain suggests that contextual information is more effective when it is presented as an explicitly ordered input sequence, rather than injected through loosely coupled concatenation. 
The improvement over MSAmba is also notable. 
Although both methods build upon Mamba-style sequence modeling, CAGMamba delivers clearly stronger classification performance, indicating that efficient sequence modeling alone is insufficient without explicit control over cross-modal information flow.

On CMU-MOSEI, CAGMamba obtains the best Acc-2 under the negative/non-negative protocol (87.08), the best F1 under both evaluation protocols (88.72/88.57), and the best Corr (0.82), while remaining competitive on Acc-7 (54.56) and MAE (0.51). 
It is worth noting that Semi-IIN slightly outperforms our model on the negative/positive Acc-2 protocol, Acc-7, and MAE, indicating that fine-grained prediction on the larger and more diverse MOSEI benchmark remains challenging. 
Nevertheless, CAGMamba provides the most balanced overall performance across classification and correlation metrics, suggesting that the proposed design improves not only decision accuracy but also the consistency of sentiment ordering.

On CH-SIMS, CAGMamba achieves the best Acc-2 (83.13), F1 (83.15), and Corr (0.703), while obtaining competitive results on Acc-3, Acc-5, and MAE. 
Compared with MSAmba, our model improves Acc-2 and F1 by 0.83 and 1.40 points, respectively. 
Compared with MMML, the gains further increase to 1.95 points on Acc-2 and 2.17 points on F1. 
Although the margins on Acc-5 and MAE are smaller, the consistent advantage on binary sentiment discrimination and correlation indicates that the proposed framework is particularly effective at capturing coarse-grained polarity and relative sentiment ordering in Chinese multimodal data.


\subsection{Ablation Study}\label{subsec_ablation}
\begin{table}[t]
\centering
\caption{Ablation study evaluating the impact of contextual modeling and fusion strategies. The best results are marked in \textbf{bold}, while the \underline{second-best} results are underlined.}
\label{tab:ablation}

\begin{tabular}{llc}
\toprule
\textbf{Experiment} & \textbf{Configuration} & \textbf{Acc-2}$\uparrow$ \\
\midrule

\multirow{4}{*}{Context} 
& MMML (no context)                & 86.31 \\
& MMML + concat                    & 87.17 \\
& GCMN + concat                    & \underline{87.31} \\
& GCMN + gated seq.\ (ours)        & \textbf{88.19} \\

\midrule

\multirow{3}{*}{Fusion} 
& Single-path (no gate)            & 86.21 \\
& Dual-path + fixed gate           & \underline{87.31} \\
& Dual-path + learnable gate (ours)& \textbf{88.19} \\

\bottomrule
\end{tabular}
\end{table}

All ablation experiments are conducted on CMU-MOSI, and the results are reported in Table~\ref{tab:ablation}.

We first study the contribution of the context modeling strategy. 
Starting from the no-context baseline, simply introducing contextual information improves Acc-2 from 86.31 to 87.17, confirming that preceding utterances provide useful disambiguating cues for the current sentiment. 
Replacing direct concatenation with the proposed ordered sequence construction further improves performance to 87.31. 
This improvement suggests that arranging context and current utterance in an explicit temporal order enables Mamba to exploit their directional dependency more effectively than feature-level concatenation.
After further introducing learnable gating, Acc-2 increases to 88.19. Overall, the full model improves by 1.88 points over the no-context variant, indicating that the gain is cumulative and arises from the interaction between structured context modeling and adaptive fusion.

For the fusion mechanism, the single-path variant yields the lowest Acc-2 (86.21), suggesting that directly collapsing multimodal information into a single fusion path tends to blur modality-specific cues. 
Introducing separate unimodal and cross-modal paths with fixed weights improves the score to 87.31, showing that preserving modality-specific representations alongside shared cross-modal features is already beneficial. 
Replacing fixed weights with learnable gates yields an additional gain of 88.19. This result indicates that the usefulness of cross-modal evidence is instance-dependent: some samples benefit from stronger cross-modal interaction, while others rely more on unimodal evidence. 
The proposed gating mechanism enables the model to adapt to this variation dynamically, rather than enforcing a single fusion ratio across all samples.
\begin{table}[t]
\centering
\caption{Ablation study on context window size on CMU-MOSI. $K_\text{t}$ and $K_\text{a}$ denote the number of preceding utterances used as context for text and audio, respectively. The best result is marked in \textbf{bold} and the second-best is \underline{underlined}.}
\label{tab:context_window}
\setlength{\tabcolsep}{4.5pt}
\begin{tabular}{l cccccccc}
\toprule
$(K_\text{t},\, K_\text{a})$ & (0,\,0) & (1,\,0) & (0,\,1) & (1,\,1) & (1,\,2) & (2,\,2) & (3,\,1) & \textbf{(2,\,1)} \\
\midrule
Acc-2$\uparrow$ & 86.33 & 86.98 & 86.72 & 87.54 & 87.15 & 87.63 & \underline{87.95} & \textbf{88.19} \\
\bottomrule
\end{tabular}
\end{table}

We further investigate the effect of context window size on CMU-MOSI by varying the number of preceding utterances $K_\text{t}$ and $K_\text{a}$ used as context for text and audio, respectively. As shown in Table~\ref{tab:context_window}, removing context entirely ($K_\text{t} = K_\text{a} = 0$) yields an Acc-2 of 86.33, confirming the importance of contextual information. Comparing single-modality context, text-only context ($K_\text{t} = 1, K_\text{a} = 0$) outperforms audio-only context ($K_\text{t} = 0, K_\text{a} = 1$), indicating that textual context provides stronger disambiguating cues. Increasing the audio window beyond one does not bring further gains, suggesting that prosodic patterns are more temporally localized than lexical content. The asymmetric configuration ($K_\text{t} = 2, K_\text{a} = 1$) achieves the best Acc-2 of 88.19, outperforming both symmetric settings and the larger variant ($K_\text{t} = 3, K_\text{a} = 1$), confirming that a moderate text-dominant context strategy is most effective.




\subsection{Efficiency Study}\label{subsec_efficiency}
\begin{table}[t]
\centering
\caption{Efficiency comparison in terms of model size (Params), computational cost (FLOPs), and performance (Acc-2).  The best results are marked in \textbf{bold}, while the \underline{second-best} results are underlined.}
\label{tab:efficiency_mosi}

\begin{tabular}{lccc}
\toprule
\textbf{Method} & \textbf{Params} & \textbf{FLOPs} & \textbf{Acc-2}$\uparrow$ \\
\midrule

Vision Mamba & 1.40M & 0.12G & 84.97/85.49 \\
Video Mamba  & \underline{1.41M} & \underline{0.13G} & 85.74/86.89 \\
Transformer  & 4.45M & 0.38G & 84.74/86.20 \\
MSAmba       & \underline{1.41M} & \underline{0.13G} & \underline{85.99}/\underline{87.43} \\

\midrule
GCMN (ours)  & \textbf{0.80M} & \textbf{0.10G} & \textbf{88.19}/\textbf{88.15} \\

\bottomrule
\end{tabular}
\end{table}

To evaluate computational efficiency, CAGMamba is compared with several lightweight sequence modeling baselines, as reported in Table ~\ref{tab:efficiency_mosi}.
CAGMamba uses only 0.80M parameters and 0.10G FLOPs, yet achieves the best Acc-2 (88.19/88.15) among all compared variants. 
Relative to the Transformer baseline, our model reduces the parameter count by approximately 82\% and FLOPs by approximately 74\%, while improving Acc-2 by 3.45 and 1.95 points under the two evaluation protocols. 
Compared with MSAmba, CAGMamba reduces parameters from 1.41M to 0.80M and FLOPs from 0.13G to 0.10G, while achieving better predictive performance. 
These results confirm that the gains stem from a more appropriate inductive bias rather than increased model capacity.



\subsection{Representation Visualization Analysis}\label{subsec_vis}
\begin{figure}[t]
  \centering
  \begin{subfigure}[b]{0.32\linewidth}
    \includegraphics[width=\linewidth]{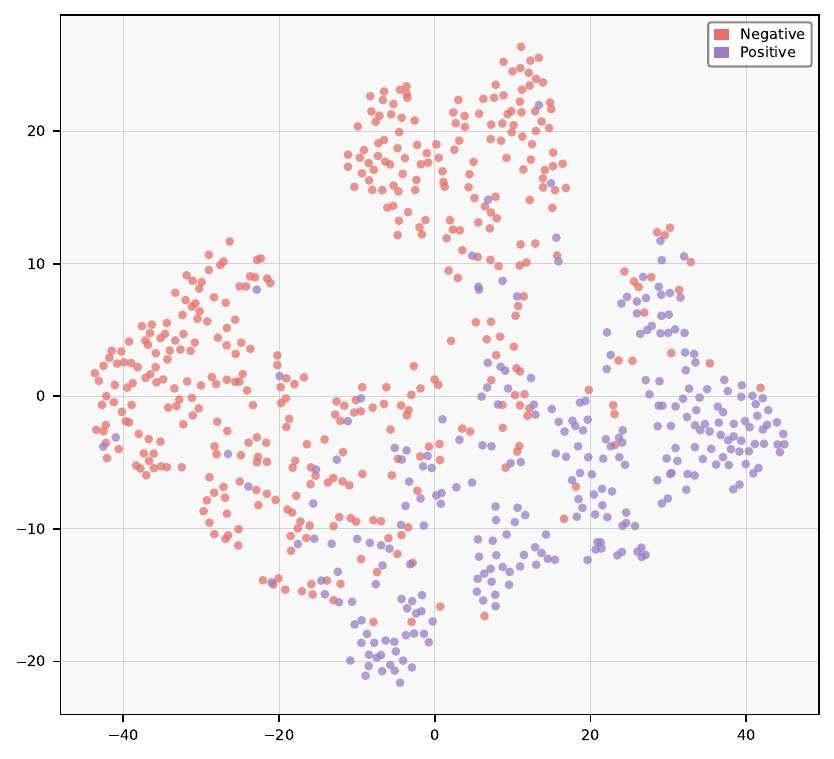}
    \caption*{(a) Cross-modal Only}
  \end{subfigure}
  \hfill
  \begin{subfigure}[b]{0.32\linewidth}
    \includegraphics[width=\linewidth]{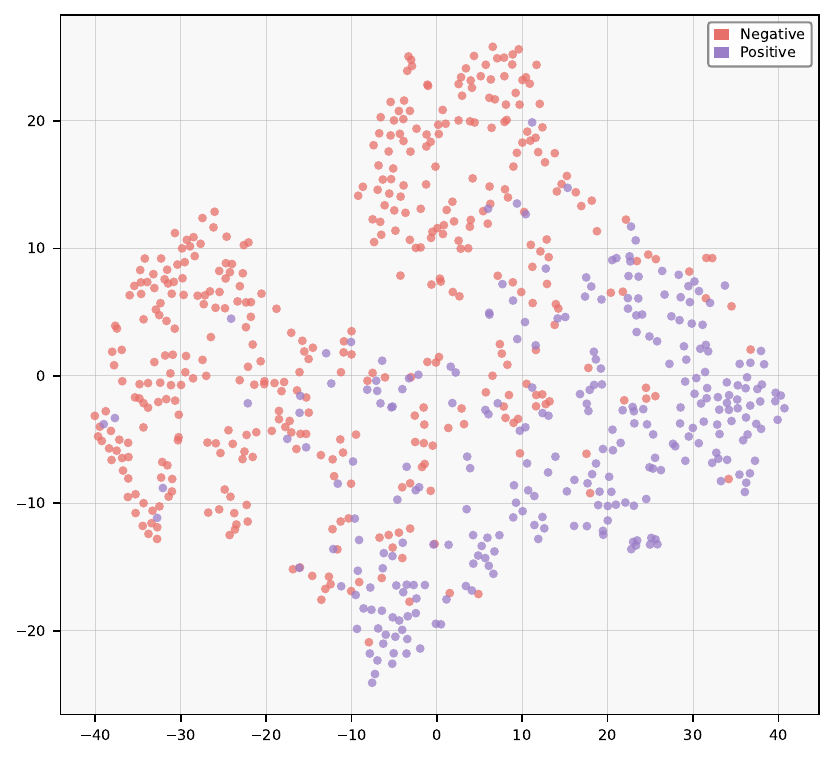}
    \caption*{(b) Fixed Fusion (0.5/0.5)}
  \end{subfigure}
  \hfill
  \begin{subfigure}[b]{0.32\linewidth}
    \includegraphics[width=\linewidth]{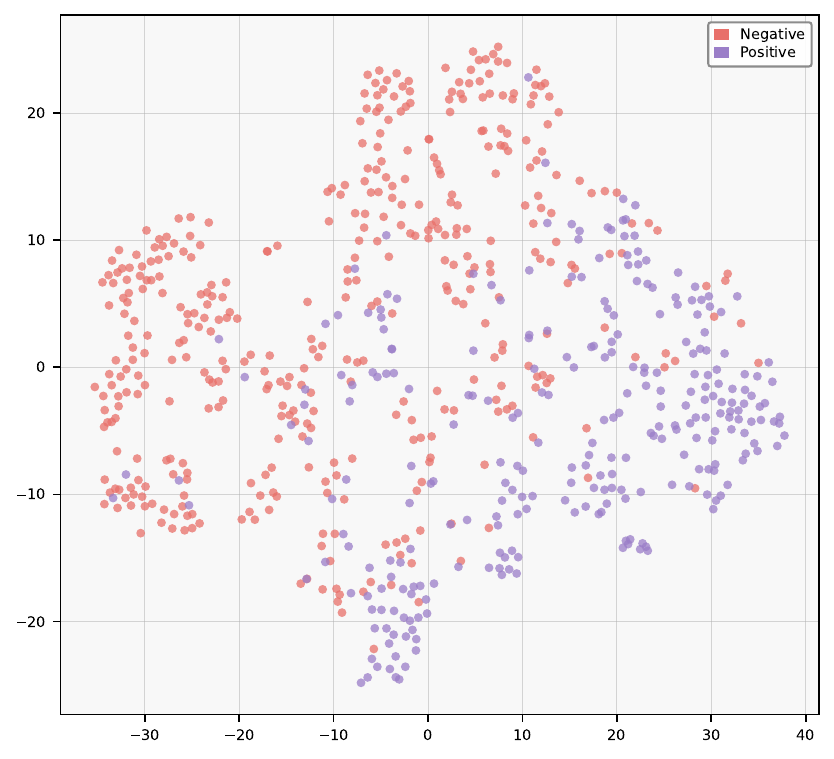}
    \caption*{(c) Gating Fusion (Ours)}
  \end{subfigure}
  \caption{T-SNE visualization of the learned representations on CMU-MOSI. Compared to baseline methods, our model produces more compact intra-class clusters and better inter-class separation, indicating improved representation quality.}
  \label{fig:tsne}
\end{figure}
As shown in Figure~\ref{fig:tsne}, the representations learned by each fusion variant are projected into two dimensions via t-SNE for qualitative comparison.
The cross-modal-only variant exhibits substantial overlap between positive and negative samples, and the fixed-fusion variant only partially alleviates this issue with a diffuse class boundary. 
In contrast, CAGMamba produces more compact intra-class clusters and clearer inter-class separation, suggesting that the proposed gating mechanism effectively suppresses redundant cross-modal signals while retaining discriminative modality-specific evidence. 
This qualitative pattern is consistent with the quantitative results: the proposed gating mechanism enables CAGMamba to improve not only classification accuracy, but also the geometric structure of the representation space learned by the model.

\section{Conclusion}
This paper presented CAGMamba, a context-aware gated cross-modal Mamba framework for dialogue-based multimodal sentiment analysis. The proposed framework organizes contextual and current-utterance features into a temporally ordered sequence, enabling Mamba's selective state space mechanism to capture directional sentiment transitions across dialogue turns. A Gated Cross-Modal Mamba Network (GCMN) is further introduced to adaptively balance cross-modal fusion and modality-specific preservation through learnable gating, coupled with a three-branch multi-task objective that jointly supervises text, audio, and fused predictions. Experiments on three benchmark datasets demonstrate state-of-the-art or competitive performance across multiple metrics, with only 0.80M parameters, substantially fewer than comparable baselines.

Despite the progress, several open challenges remain in dialogue-based multimodal sentiment analysis, including the limited scale and domain diversity of existing benchmarks and the reliance on utterance-level scalar annotations that may oversimplify nuanced affective expressions. Addressing these broader issues will be important for advancing more generalizable multimodal sentiment systems.



\bibliographystyle{splncs04_unsrt}
\bibliography{reference}

\end{document}